\documentclass[10pt,journal,cspaper,compsoc]{IEEEtran}
\ifCLASSOPTIONcompsoc
\else
\fi
\ifCLASSINFOpdf
\else
\fi
\usepackage{fixltx2e}
\usepackage[hyphens]{url}
\usepackage{tikz}
\usepackage{textcomp}
\usepackage{hyperref}
\usepackage{subscript}

\usepackage{mathtools}  % MARCO AGGIUNTO
\usepackage{multirow} % MARCO AGGIUNTO
\usepackage{booktabs} % LORENZO AGGIUNTO
\usepackage[english,norounding]{rccol} % LORENZO AGGIUNTO

\usepackage{fancyhdr}
 
\begin{document}
\title{SentiWords: Deriving a High Precision and High Coverage Lexicon for Sentiment Analysis}

\author{Lorenzo~Gatti, %~\IEEEmembership{Fellow,~OSA,}
        Marco~Guerini, %~\IEEEmembership{Life~Fellow,~IEEE}% <-this % stops a space
        and~Marco~Turchi%~\IEEEmembership{Life~Fellow,~IEEE}% <-this % stops a space
\IEEEcompsocitemizethanks{
\IEEEcompsocthanksitem L. Gatti, M. Guerini and M. Turchi are with FBK, Trento, % Via Sommarive 18, 38123 Povo (TN),
Italy.\protect\\
E-mail: \{l.gatti, guerini, turchi\}@fbk.eu.
%\IEEEcompsocthanksitem M. Guerini is with Trento RISE, Trento, %Via Sommarive 18, 38123 Povo (TN),
%Italy.\protect\\
%E-mail: marco.guerini@trentorise.eu.
}% <-this % stops a space
\thanks{}}        
\IEEEcompsoctitleabstractindextext{%
\begin{abstract}
Deriving prior polarity lexica for sentiment analysis --  where positive or negative scores are associated with words out of context -- is a challenging task. Usually, a trade-off between precision and coverage is hard to find, and it depends on the methodology used to build the lexicon. Manually annotated lexica provide a high precision but lack in coverage, whereas automatic derivation from pre-existing knowledge guarantees high coverage at the cost of a lower precision.
Since the automatic derivation of prior polarities is less time consuming than manual annotation, there has been a great bloom of these approaches, in particular based on the SentiWordNet resource. In this paper, we compare the most frequently used techniques based on SentiWordNet with newer ones and blend them in a learning framework (a so called `ensemble method').
By taking advantage of manually built prior polarity lexica, our ensemble method is better able to predict the prior value of unseen words and to outperform all the other SentiWordNet approaches.
Using this technique we have built \emph{SentiWords}, a prior polarity lexicon of approximately 155,000 words, 
that has both a high precision and a high coverage. We finally show that in sentiment analysis tasks, using our lexicon allows us to  outperform both the single metrics derived from SentiWordNet and popular manually annotated sentiment lexica.
\end{abstract}

% not want either math or citations to appear in the abstract.

% Note that keywords are not normally used for peer review papers.
}

\maketitle
 \enlargethispage{-0.8cm}

% is not selected <OR> if conference mode is selected - because compsoc
% conference papers position the abstract like regular (non-compsoc)
\IEEEdisplaynotcompsoctitleabstractindextext

 \IEEEoverridecommandlockouts
% \IEEEpubid{\makebox[\columnwidth]{978-1-4799-7492-4/15/\$31.00~
% \copyright2015
% IEEE \hfill} \hspace{\columnsep}\makebox[\columnwidth]{ }} 

\newcommand\copyrighttext{%
  \footnotesize \textcopyright 2015 IEEE. Personal use of this material is permitted. Permission from IEEE must be obtained for all other uses, in any current or future media, including reprinting/republishing this material for advertising or promotional purposes, creating new collective works, for resale or redistribution to servers or lists, or reuse of any copyrighted component of this work in other works.\\* DOI: \href{http://ieeexplore.ieee.org/xpl/articleDetails.jsp?reload=true&arnumber=7239537}{10.1109/TAFFC.2015.2476456}}
\newcommand\copyrightnotice{%
\begin{tikzpicture}[remember picture,overlay]
\node[anchor=south,yshift=10pt] at (current page.south) {\fbox{\parbox{\dimexpr\textwidth-\fboxsep-\fboxrule\relax}{\copyrighttext}}};
\end{tikzpicture}%
}

% \fi
\maketitle
\copyrightnotice

\section{Introduction}

{\PARstart{I}{n} sentiment analysis many approaches employ specialized lexica -- i.e. lists of positive and negative words -- often in conjunction with other methods  (usually machine learning based)  \cite{liu2012survey}, to assign sentiment scores to texts. }
In most of these lexica, words are associated with their prior polarity, i.e. if that word out of context evokes something positive or something negative. For example, \emph{wonderful} has a positive connotation -- prior polarity -- while \emph{horrible} has a negative one. These approaches, based on prior polarity lexica, {are so popular because they do not need} 
word sense disambiguation to assign an affective score to a word, and they are often largely domain-independent. 
{Prior polarity lexica can be roughly divided into two groups: those that are manually built (either hiring expert annotators such as linguists or by crowdsourcing the annotation on web platforms such as Mechanical Turk), and those that are automatically derived from pre-existing knowledge. While the first kind of lexica has a high precision but a low coverage, the opposite holds for the second kind. } 

{In this paper, we aim to understand if blending both approaches we can build a lexicon that has both a high coverage and a high precision.} We focus on SentiWordNet (henceforth SWN), a resource that has been widely adopted since it provides a broad-coverage lexicon -- built in a semi-automatic manner -- for English \cite{Esuli06}. 
Given that SWN provides polarity scores for each word sense (also called `posterior polarities'), it is necessary to derive prior polarities from the posteriors. 

Several formulae to compute prior polarities starting from posterior polarity scores have been proposed in the literature.  {By comparing  the formulae against manually built prior polarity lexica we show that some of these formulae are better than others at estimating prior polarities} and can represent a fairer state-of-the-art approach using SWN. On top of this, we attempt to outperform the state-of-the-art formula using an `ensemble' learning framework that combines the various formulae together {and takes advantage of manually built prior polarity lexica to better predict the value of unseen words. In this way we construct a sentiment lexicon that has both a high coverage and a high precision.}

In detail,  the first part of the paper -- that is based on our previous work, presented in \cite{guerini2013sentiment} -- addresses three main research questions about words prior polarity computation:  
\enlargethispage{-1.2cm}
(\emph{i}) is there any relevant difference in the posterior-to-prior polarity formulae performance (both in regression and classification tasks)?
(\emph{ii}) Is there any relevant variation in prior polarity values if we use different releases of SWN (i.e. SWN\textsubscript{1} or SWN\textsubscript{3})? 
(\emph{iii}) Can a learning framework boost the performance of such formulae?

In the second part of the paper --  that represents the novel contribution of the present work -- we introduce \emph{SentiWords}\footnote{The resource can be downloaded at \url{https://hlt.fbk.eu/technologies/sentiwords}.}, a prior polarity lexicon produced according to the lesson learned from the first part of the paper, and we answer an additional set of questions regarding sentiment analysis of sentences using words prior polarities: 
(\emph{i}) does \emph{SentiWords} still have better performance compared to the posterior-to-prior polarity formulae? 
(\emph{ii}) How important is the coverage of the lexicon compared to other handmade lexica (more precise but smaller)?
(\emph{iii}) How well does \emph{SentiWords} perform across datasets compared to a specialized posterior-polarities lexicon? 

In the following two sections, we present a series of experiments, both in regression and classification tasks, that give an answer to the aforementioned research questions. The results support the hypothesis that using a learning framework can improve on the state-of-the-art performance in posterior-to-prior computation and that using \emph{SentiWords} in sentiment analysis provides better results than other available lexica. 

\section{Related Work}
\label{sec:Related_Work}

The quest for a high precision and high coverage lexicon, where words are associated with either sentiment or emotion scores, has several reasons.
First, it is fundamental for tasks such as affective modification of existing texts, where words polarity together with their scores are necessary {for creating multiple versions of a text, varying its affective dimension} \cite{inkpen2006generating,Guerini2008,Whitehead10}.

{Second, while in sentiment analysis compositionality (i.e. methods to compute the score of a sentence by combining the scores of the words in its syntactic tree)  plays a crucial role, list of words associated with their sentiment score are still a fundamental prerequisite for this task.}
Works using compositional approaches worth mentioning  are: \cite{socher2013recursive}, that uses recursive neural networks to learn compositional rules for sentiment analysis, while \cite{polanyi2006contextual,moilanen2007sentiment,neviarouskaya2011affect} exploit hand-coded rules. In this respect, compositional approaches represent a promising new trend, since all other approaches, either using semantic similarity or Bag-of-Words (BOW) based machine-learning, cannot handle, for example, cases of texts with the same wording but different word order:
``\emph{The dangerous killer escaped one month ago, but lately he was arrested}" (positive)
 vs. ``\emph{The dangerous killer was arrested one month ago, but lately he escaped}" (negative).
The work in ~\cite{wangbaselines} partially accounts for this problem arguing that using word bigrams allows improvement over BOW based methods, where words are taken as features in isolation. This way it is possible to capture simple compositional phenomena such as polarity reversing in ``\emph{killing cancer}".

Finally, tasks such as copywriting, where evocative names are a key element to a successful product \cite{ozbalcomputational,ozbal2012brand} require exhaustive lists of emotion 
related words. In such cases no context is given and the brand name alone, with its perceived prior polarity, is responsible for stating the area of competition and evoking 
semantic associations. 
Evoking emotions is also fundamental for a successful name: consider names of a perfume such as 
\emph{Obsession}, or technological products such as MacBook \emph{Air}.

We now provide a  review focusing on research efforts put towards building sentiment and emotion lexica, regardless of the approach in which 
such lists are then used (machine learning, rule based or deep learning). A general overview can be found in~\cite{pang2008opinion,liu2012survey,wilson:AAAI-04,Paltoglou2010}.

\textbf{Sentiment Lexica}.
In recent years there has been an increasing focus on producing lists of words with affective polarities, to be used in sentiment analysis. When building such lists, a trade-off between coverage and precision of the resource has to be found.  {{The highest precision is obtained with manually annotated lexica, but these are usually smaller due to the time and costs associated with the annotation task. Automatically created resources are usually larger, but their precision is highly dependent on the annotation algorithm \cite{Heerschop:2011aa} and, in general, not {as accurate as} manual resources.}}

One of the most well-known resources is~\emph{SentiWordNet} (SWN)~\cite{Esuli06,baccianella2010sentiwordnet}, in which each entry is a set of \texttt{lemma\#PoS\#sense-number} sharing the same meaning, called \emph{synset}.  Starting from SWN, several prior polarities for words in the form \texttt{lemma\#PoS}, can be computed (e.g. considering only the first-sense or averaging on all the senses). 
These approaches, detailed in \cite{guerini2013sentiment}, produce a list of approximately 155,000 words, where the lower precision given by the automatic scoring  of SWN is compensated by the high coverage. SWN and formulae for prior computation will be thoroughly described in Section \ref{sec:approach}.

Another widely used resource is the \emph{Affective Norms for English Words} (ANEW) \cite{bradley1999affective}, providing valence scores for roughly 1,000 words, which were manually assigned by several annotators. This resource has a low coverage, but the precision is very high. 
Similarly, the \emph{SO-CAL} entries \cite{taboada2011lexicon}  consist of roughly 4,000 words manually tagged by a small number of {linguists} with a multi-class label (from \texttt{very\_negative} to \texttt{very\_positive}). These ratings were further validated through crowdsourcing. 
{The Dictionary of Affect in Language (DAL) contains roughly 9,000 words manually rated along the dimensions `pleasantness', `activation' and `imagery' \cite{whissell1989dictionary}.}
More recently, a resource replicating the ANEW annotation approach using crowdsourcing was released by Warriner and colleagues~\cite{Warriner2013norms}, providing sentiment scores for approximately 14,000 words (this lexicon will be referred to as \emph{Warr} henceforth).
Interestingly, this resource includes the most frequently used English words, so -- even if its coverage is still far lower than SWN --  it grants a high coverage, with human precision, of language use. Finally,  the \emph{General Inquirer} lexicon \cite{stone1966general} provides a binary classification  (\texttt{positive}/\texttt{negative}) of approximately 4,000 sentiment-bearing words manually annotated, while the resource presented in \cite{wilson2005recognizing} expands the General Inquirer to 6,000 words. 

\textbf{Emotion Lexica}.
One of the most used resources is \emph{WordNetAffect}~\cite{strappaLREC04} which contains manually assigned affective labels to WordNet synsets 
(\textsc{anger}, \textsc{joy}, \textsc{fear}, etc.). It currently provides 900 annotated synsets and 1,600 words in the form \texttt{lemma\#PoS\#sense}, corresponding to roughly 1,000 lemma-PoS.  \emph{AffectNet}, part of the SenticNet project \cite{cambria2012sentic}, contains approximately 10,000 words (out of 23,000 entries) taken from ConceptNet and aligned with WordNetAffect. This 
resource extends WordNetAffect labels to concepts such as `have breakfast'.
\emph{Fuzzy Affect Lexicon} \cite{subasic2001affect} contains roughly 4,000 lemma-PoS manually annotated by one linguist using 80 emotion labels. \emph{EmoLex} \cite{mohammad2013crowdsourcing} contains almost 10,000 lemmas annotated with an intensity label for each emotion using Mechanical Turk.
Finally~\emph{Affect database} is an extension of SentiFul \cite{Neviarouskaya:2007fk} and contains 2,500 words in the form \texttt{lemma\#PoS}, while \emph{DepecheMood} \cite{staiano2014depechemood} contains about 37,000 words also in the \texttt{lemma\#PoS} format, and was automatically built by harvesting crowd-sourced affective annotation from a social news network. These latter two lexica are the only ones providing words annotated with emotion scores, rather than just with labels.

\section{Proposed Approach}
\label{sec:approach}

In the broad field of Sentiment Analysis we will {first} focus on the specific problem of {words} posterior-to-prior polarity assessment, using SWN both in regression and classification experiments. 

For the regression task, we tackle the problem of assigning affective scores (along a continuum between -1 and 1) to words, using posterior-to-prior polarity formulae. For the classification task (assessing whether a word is either \emph{positive} or \emph{negative}) we use the same formulae, but considering just the sign of the result.  In these experiments we also use {an ensemble method} which combines the various formulae together. The underlying hypothesis is that by blending these formulae, and looking at the same information from different perspectives (i.e. the posterior polarities provided by SWN combined in various ways), we can obtain a better prediction.

{In the second part of the paper we will validate the improvement we can obtain in a simple sentiment analysis task with the lexicon produced by our ensemble method over the single SWN metrics and over other widely used handmade lexica. To this end, we run an extensive series of experiments on two different datasets of sentences that represents different forms of language use, i.e. news headlines, with simplified syntax and lexicon, and sentences extracted from movie reviews, with normal language use. Also in this case, we face both regression and classification tasks.}

\subsection{SentiWordNet}
\label{sec:SWN}

SentiWordNet \cite{Esuli06} is a lexical resource composed of ``synsets",  i.e. sets of 
{\texttt{lemma\#PoS\#sense-number} tuples (where the smallest sense-number corresponds to the most frequent sense of the lemma)
sharing the same meaning. Each synset \texttt{s} is associated with the numerical scores \texttt{Pos(s)} and \texttt{Neg(s)}, which range from 0 to 1. These scores represent the positive and negative valence (or posterior  polarity) of the synset, and are shared by each entry in the synset. 
    The scores were automatically assigned by a classifier committee trained on the glosses of three subsets of WordNet: one composed of positive synsets, one of negative synsets and one containing ``neutral'' synsets, i.e. synsets that are neither positive nor negative.  The positive and negative subsets were constructed by (\emph{i}) finding the synsets containing 14 ``paradigmatic'' positive and negative words (e.g. \texttt{good\#a\#1}), and (\emph{ii}) automatically expanded by traversing the WordNet hierarchy to find ``related" synsets, using the method described in \cite{strappaLREC04}. Neutral synsets are those that do not belong to the other two subsets and that do not contain terms marked as Positive or Negative in the General Inquirer lexicon. 
}

Obviously, different senses of a \texttt{lemma\#PoS} can have different polarities. In Table \ref{tab:sentiwncoldsenses}, the first 5 senses of \texttt{cold\#a} present all  possible combinations, including mixed scores (\texttt{cold\#a\#4}), where positive and negative valences are assigned to the same sense. Intuitively, mixed scores for the same sense are acceptable, as in ``\emph{cold} beer'' (positive) vs. ``\emph{cold} pizza'' (negative).

\begin{table} [ht]
	\caption{First five \textit{SentiWordNet} entries for \texttt{cold\#a}}
	\label{tab:sentiwncoldsenses}
	\centering
	{
		\begin{tabular}{lrrr}
		 \toprule
      Synset\_ID & \multicolumn{1}{c}{Pos(s)} & \multicolumn{1}{c}{Neg(s)} & SynsetTerms\\
\midrule      
      1207406&0.0&0.75&\texttt{cold\#a\#1}\\
      1212558&0.0&0.75&\texttt{cold\#a\#2}\\
      1024433&0.0&0.0&\texttt{cold\#a\#3}\\
      2443231&0.125&0.375&\texttt{cold\#a\#4}\\
      1695706&0.625&0.0&\texttt{cold\#a\#5}\\
 \bottomrule
		\end{tabular}
		}
\end{table} 

In our experiments we use two different versions of SWN: SentiWordNet 1.0 (SWN\textsubscript{1}), the first release of SWN, and its updated version SentiWordNet 3.0 \cite{baccianella2010sentiwordnet} (SWN\textsubscript{3}).
{The latter differs from the former because 
(\emph{i}) it annotates WordNet 3.0 instead of WordNet 2.0;
(\emph{ii}) it ``corrects'' the classifiers scores with a random-walk process, where the glosses are used to adjust the negativity and positivity scores of the synsets
(\emph{iii}) it uses different, manually annotated glosses, both for training the classifiers and for the previous step.
This new annotation algorithm led to an increase in the accuracy of posterior polarities over the first version, as reported by the authors.}

\subsection{Prior Polarities Formulae}
\label{sec:PriorF}

In this section, we review the strategies for computing prior polarities from SWN used in previous studies. All the proposed approaches try to estimate the prior polarity from the posterior polarities of all the senses for a single lemma-PoS. Given a lemma-PoS with $n$ senses (\texttt{lemma\#PoS\#n}), every formula $f$ is independently applied to {$posScore$ and $negScore$ (which are the ordered sets of all the \texttt{Pos(s)} and all the \texttt{Neg(s)} for that lemma-PoS, respectively).} This produces two scores {in the range $[0, 1]$}, $f(posScore)$ and $f(negScore)$, for each lemma-PoS. To obtain a unique prior polarity, $f(posScore)$ and $f(negScore)$ can be mapped according to different strategies:
\begin{footnotesize}
\begin{align*}
f_m &=
\begin{cases}
\hphantom{-}f(posScore) & \text{if  $f(posScore)\geq f(negScore)$}\\
-f(negScore) & \text{otherwise}
\end{cases}\\[5pt]
f_d &= f(posScore) - f(negScore)
\end{align*}
\end{footnotesize}
where $f_m$ computes the absolute maximum of the two scores, while $f_d$ computes the difference between them. {Both numbers are in the range $[-1, 1]$.} 
So, considering the first 5 senses of \texttt{cold\#a} in Table \ref{tab:sentiwncoldsenses}, $f(posScore)$ will be derived from the \texttt{Pos(s)}  values {\small ${<}0.0 , 0.0 , 0.0 , 0.125 , 0.625{>}$}, while $f(negScore)$ from {\small ${<}0.750 , 0.750 , 0.0 , 0.375 , 0.0{>}$}. Then, the final polarity strength will be either $f_m$ or $f_d$. The formulae ($f$) we tested are the following:

\textbf{fs}. In this formula only the first (and thus most frequent) sense is considered for the given \texttt{lemma\#PoS}. This is equivalent to considering just the SWN score for \texttt{lemma\#PoS\#1}. Based on \cite{10.1109/T-AFFC.2011.1,agrawal2009using,Guerini2008,chowdhury-EtAl:2013:SemEval-2013}, %(that uses the $fs_m$ approach), 
this is the most basic form of prior polarities.

\textbf{mean}. It calculates the mean of the positive and negative scores for all the senses of the given \texttt{lemma\#PoS}. It  was used in  \cite{thet2009sentiment,denecke2009sentiwordnet,devitt2007sentiment,sing2012development}.

\textbf{uni}. Based on \cite{10.1109/T-AFFC.2011.1}, it considers only senses having a \texttt{Pos(s)} greater than or equal to the corresponding \texttt{Neg(s)}, and greater than 0 (the $stronglyPos$ set). In the case where $posScore$ is equal to $negScore$ {(thus also $f(posScore)=f(negScore)$)}, the one with the highest weight is returned, where weights are defined as the cardinality of $stronglyPos$ divided by the total number of senses. The same applies for the negative senses. This is the only method, together with $rnd$, for which we cannot apply $f_d$, as it returns a positive or negative score according to the weight.

\textbf{uniw}. Like $uni$ but without the weighting system.

\textbf{w1}. This formula weights each sense with a geometric series of ratio 1/2. The rationale behind this choice is based on the assumption that more frequent senses should bear more ``affective weight'' than rare senses when computing the prior polarity of a word.  The system presented in \cite{chaumartin2007upar7} uses a similar approach of weighted mean.

\textbf{w2}. Similar to the $w1$, this formula weigths each lemma with a harmonic series, see for example \cite{Berardi201338} (where $w2$ appears with the $f_d$ variant). %and  \cite{denecke2008accessing}.

%%\medskip
On top of these formulae, we implemented some new formulae that were relevant to our task and, to our knowledge, have not been proposed in the literature. 
These formulae mimic those discussed previously, but they are built under a different assumption: that the saliency of a word prior polarity might be more related to its posterior  scores, rather than to sense frequencies.
Thus we ordered $posScore$ and $negScore$ by strength, giving more relevance to ``strongly valenced" senses.
For instance, in Table \ref{tab:sentiwncoldsenses}, $posScore$ and $negScore$ for \texttt{cold\#a} become {\small ${<}0.625 , 0.125 , 0.0 , 0.0 , 0.0{>}$} and {\small ${<}0.750 , 0.750 , 0.375 , 0.0 , 0.0{>}$} respectively.

\textbf{w1s} and \textbf{w2s}. These are similar to $w1$ and $w2$, but senses are ordered by strength (sorting \texttt{Pos(s)} and \texttt{Neg(s)} independently).

\textbf{w1n} and \textbf{w2n}. The same as $w1$ and $w2$ respectively, but without considering senses that have a 0 score for both \texttt{Pos(s)} and \texttt{Neg(s)}. Our motivation is that null senses constitute noise for the purposes of lexicon bootstrapping.

\textbf{w1sn} and \textbf{w2sn}. The same as $w1s$ and $w2s$, but without considering senses that have a 0 score for both \texttt{Pos(s)} and \texttt{Neg(s)} respectively.

\textbf{median}. Returns the median of the senses ordered by polarity score.

{\textbf{max}. Returns $max(posScore)$ and $max(negScore)$, i.e. it returns the highest positive and negative values among all senses.}

All these prior polarities formulae are compared to two gold
standards sentiment lexica (one for regression, one for classification) both separately, as in the works mentioned above,  and combined together in a
learning framework (to see whether combining these features -- that
capture different aspect of prior polarities --  can further improve
the results).

Finally, we implemented two variants of a prior polarity random baseline to assess possible advantages of approaches using SWN:

\textbf{rnd.} This formula represents the basic baseline random
approach. It  simply returns a random number between -1 and 1 for any
given \texttt{lemma\#PoS}.

\textbf{swnrnd.} This is an advanced random approach
that incorporates some ``knowledge'' from SWN. It takes the scores of a random sense for the given \texttt{lemma\#PoS}. We believe this is a fairer baseline than $rnd$ since SWN information can possibly constrain the values. A similar approach has been used in  \cite{qu2008sentence}. 

{\textbf{majority\_class.} For the classification experiments we considered an additional baseline that always outputs the class with the higher number of instances, to account for imbalanced datasets.}

\subsection{Learning Algorithms}
\label{sec:ML}
{
All the proposed formulae try to estimate the prior polarity score from the posterior polarities of all the senses for a single lemma-PoS. Each formula has its own partial view of all the information available in the senses,
 and different formulae can identify complementary information, e.g. some consider only the first sense (\textit{fs}), others only the highest positive and negative values among all senses (\textit{max}).
An extension to the use of each formula in isolation consists in taking all the predicted scores produced by each formula and defining ensemble methods that, given the formulae prior polarity predictions, fuse them and emit a unique prior polarity. 
} 

{The most used  ensemble method is the majority voting schema, that assigns to an unseen lemma-PoS the label with the highest number of votes received from the formulae. While it is quite straightforward for classification problems (see \cite{rokach2010pattern}, chapter 3), combining regression scores can require \textit{ad-hoc} decisions. To propose a solution that can be easily applied to both regression and classification, we take advantage of the classic fusion learning framework, where a regressor/classifier is fed with the output of several regressors/classifiers (in our context these are the formulae outputs) and learns from the training data the optimal way to combine them into a single score (prior polarity).}

{For this purpose, we used two non-parametric learning approaches, Support Vector Machines (SVMs) \cite{shawe2004kernel} and Gaussian Processes (GPs) \cite{2006gaussian}, to test the performance of all the metrics in conjunction.  SVMs are non-parametric deterministic algorithms that have been widely used in several fields. GPs, on the other hand, are an extremely flexible non-parametric probabilistic framework able to explicitly model uncertainty, that only recently have been receiving increased attention in the NLP community. An exhaustive explanation of the two methodologies can be found in \cite{shawe2004kernel,mammone2009support}  and \cite{2006gaussian}.}

{In the SVM experiments, we use $C$-SVM and $\epsilon$-SVM implemented in the LIBSVM toolbox \cite{CC01a}. The selection of the kernel (linear, polynomial, radial basis function and sigmoid)  and the optimization of the parameters are carried out through grid search in 10-fold cross-validation.
As demonstrated in \cite{weston2000feature}, SVMs can benefit from the application of feature selection techniques. For this purpose, Randomized Lasso, or stability selection \cite{Meinshausen2010} is applied before training the SVM learner. In our experiments we set the fraction of the data to be sampled at each iteration to 75\%, the selection threshold to 25\% and the number of re-samples to 1,000. We refer to %this experiments 
these as \textit{SVMfs}.}

{GP\footnote{More details on the differences between GPs for regression and classification and the GP kernels are available in  \S2, \S3, \S4 in \cite{2006gaussian}} regression models with Gaussian noise are a rare exception where the exact inference with likelihood functions is tractable. Unfortunately, this is not valid for the classification task where an approximation method (Laplace \cite{williams1998bayesian} in our experiments) is required.
Different kernels are tested (covariance for constant functions, linear with and without automatic relevance determination (ARD)\footnote{$linone$ and $linard$ in the result tables, respectively.}, Matern, neural network, etc.) and the linear logistic ($lll$) and probit regression ($prl$) likelihood functions are evaluated in classification.}  All the GP models were implemented using the GPML Matlab toolbox, and the  optimization of kernel parameters is performed iteratively maximizing the marginal likelihood (or in classification, the Laplace approximation of the marginal likelihood).  The maximum number of iterations was set to 100.
{ A property of GPs is their capability of weighting the features differently according to their importance in the data. 
This is referred to as the automatic relevance determination kernel (ARD). }

\section{Human-annotated Sentiment Lexica}
\label{sec:GOLD}

To assess how well prior polarity formulae perform, a gold standard with word polarities provided by human annotators is needed.
In the following we describe in detail the two resources we used for our experiments, namely ANEW for the regression experiments and the General Inquirer for the classification.

\subsection{ANEW}
\label{sec:AN}
ANEW \cite{bradley1999affective} is a resource developed to provide a set of normative emotional ratings for a large number of words (roughly 1,000{, half of them taken from similar previous experiments \cite{mehrabian1974approach,bellezza1986words}}) in the English language. It contains a set of words that have been rated in terms of pleasure (affective valence), arousal, and dominance.
{The ratings were collected from students, divided in groups balanced for gender, using the ``Self-Assessment Manikin'', an affective rating system that uses graphic representations to depict  values (e.g. happy/unhappy, excited/calm, controlled/in-control) along different emotional dimensions. Students were asked to select which image represents how they felt when reading each word.
Words were shown in different order between the groups, and they were presented in isolation (i.e. no context was provided). This means that }this resource represents a human validation of prior polarity scores for the given words, and can be used as a gold standard.
For each word ANEW provides two main metrics: $anew_\mu$, which correspond to the average score of the annotators, and $anew_\sigma$, which gives the variance in annotators scores for the given word.
{For our task we only considered the valence rating, i.e. the degree of positivity or negativity of a word}.

\subsection{General Inquirer}
\label{sec:GI}

The Harvard General Inquirer dictionary (henceforth GI) is a widely used resource, built for automatic text analysis \cite{stone1966general}. Its latest
revision\footnote{\url{www.wjh.harvard.edu/~inquirer/}} contains 11,789 words, tagged with 182 semantic and pragmatic labels, as well as with their part of speech. Words and their categories were initially taken from the Harvard IV-4 Psychosociological Dictionary \cite{harvardiv} and the Lasswell Value Dictionary
\cite{lasswell1969lasswell}. {The GI categories were defined to be used in social-science content-analysis research applications, but this resource has extensively been used for sentiment analysis too.}
For this paper we consider the \texttt{Positive} and \texttt{Negative} categories (1,915 words and 2,291 words respectively, for a total of 4,206 affective words), {which indicate words with a positive or negative valence. As with ANEW, since these words do not have a context, we consider the labels as binary human-assigned prior polarities, thus suitable to be used as a gold standard.}

\section{Prior Polarities Experiments}
\label{sec:exp}

In order to use the ANEW dataset to measure the performance of  prior polarities formulae, we had to assign a PoS to all the words to obtain the SWN \texttt{lemma\#PoS} format. To do
so, we proceeded as follows: for each word, check if it is present among both SWN\textsubscript{1} and SWN\textsubscript{3} lemmas; if not, lemmatize the word with the TextPro tool suite
\cite{pianta2008textpro} and check if the lemma is present instead\footnote{We did not lemmatize everything to avoid duplications (for example, if we lemmatize the ANEW entry \emph{addicted}, we obtain \emph{addict}, which is already present in ANEW).}. If it is not found (i.e., the word cannot be aligned automatically), remove the word from
the list (this was the case for 30 words of the 1,034 present in ANEW). 
The remaining 1,004 lemmas were then associated with all the PoS present in SWN to get the final \texttt{lemma\#PoS}. 
Note that a lemma can have more than one PoS,  for example, \emph{writer} is present only as a noun (\texttt{writer\#n}), while \emph{yellow} is present as a verb, a noun and an adjective (\texttt{yellow\#v}, \texttt{yellow\#n}, \texttt{yellow\#a}). This gave us a list of 1,484 words in the \texttt{lemma\#PoS} format. 

In a similar way we
pre-processed the GI words that uses the generic \texttt{modif} label to indicate either adjective or adverb
(noun  and verb PoS were consistently used instead).
Finally, all the sense-disambiguated words in the \texttt{lemma\#PoS\#n} format were discarded (1,114
words out of the 4,206 words with positive or negative valence). 

After the two datasets were pre-processed this way, we removed the words for which the $posScore$ and $negScore$ contained all 0 in both SWN\textsubscript{1} and SWN\textsubscript{3} (523 lemma-PoS for ANEW and  484 for the
GI dataset), since these words are not informative for our experiments. The final dataset included 961 entries for ANEW and
2,557 for GI. For each lemma-PoS in GI and ANEW, we then applied the prior polarity formulae described in Section \ref{sec:PriorF}, using both SWN\textsubscript{1} and SWN\textsubscript{3} and annotated the results.

According to the nature of the human labels (real numbers or -1/1), we ran several regression and classification experiments. In both cases, each dataset was randomly split into 70\% for training and the remaining for test. This process was repeated 5 times to generate different splits. For each partition, optimization of the learning algorithm parameters was performed on the training data (in 10-fold cross-validation for SVMs). Training and test sets were normalized using z-scores.

To evaluate the performance of our regression experiments on ANEW we used the Mean Absolute Error (MAE) {and Pearson correlation coefficient.} 
Accuracy {and Cohen's kappa} were used for the classification experiments on GI instead. We opted for accuracy -- rather than F1 -- since for us True Negatives have the same importance as True Positives. For each experiment we reported the average performance and the standard deviation over the 5 random splits. In the following sections, we used Student's t-test to check if there were statistically significant differences in the results of regression experiments. An approximate randomization test \cite{yeh2000more} was used for the  classification experiments instead.

\begin{table} [hb!]
	\caption{MAE results for regression using SWN\textsubscript{1}}
	\label{tab:ANEW_SWN_1}
	\centering
	{
		\begin{tabular}{@{}lrrrr@{}}
 \toprule
Approach  & MAE$_\mu$ & MAE$_\sigma$ & $\rho_\mu$ & $\rho_\sigma$ \\
\midrule
 $rnd$ & 0.652 & 0.026 & -0.002 & 0.123\\ 
$swnrnd_m$ & 0.427 & 0.011 & 0.350 & 0.041\\
$swnrnd_d$ & 0.426 & 0.009 & 0.354 & 0.015\\
\midrule
$uniw_m$ & 0.420 & 0.009  & 0.362 & 0.035\\
$max_m$ & 0.419 & 0.009  & 0.407 & 0.027\\
$fs_d$ & 0.413 & 0.011  & 0.404 & 0.031\\
$fs_m$ & 0.412 & 0.009  & 0.393 & 0.028\\ 
$uni$ & 0.410 & 0.010  & 0.372 & 0.044\\
$uniw_d$ & 0.406 & 0.007  & 0.392 & 0.037\\
$w1sn_m$ & 0.405 & 0.011  & 0.415 & 0.033\\
$max_d$ & 0.404 & 0.005  & 0.422 & 0.036\\
$w2sn_m$ & 0.402 & 0.011  & 0.415 & 0.033\\
$median_d$ & 0.401 & 0.014  & 0.430 & 0.029\\
$w1_d$ & 0.401 & 0.010  & 0.443 & 0.034\\
$w1n_d$ & 0.399 & 0.008  & 0.428 & 0.034\\
$mean_d$ & 0.398 & 0.010  & 0.445 & 0.034\\
$w2_d$ & 0.398 & 0.010  & 0.449 & 0.034\\
$median_m$ & 0.397 & 0.015  & 0.423 & 0.031\\
$w1sn_d$ & 0.397 & 0.008  & 0.428 & 0.034\\
$w2sn_d$ & 0.397 & 0.008  & 0.428 & 0.034\\
$w2n_d$ & 0.397 & 0.008  & 0.431 & 0.034\\
$w1s_m$ & 0.396 & 0.010  & 0.431 & 0.034\\
$w1_m$ & 0.396 & 0.010  & 0.438 & 0.034\\
$w1n_m$ & 0.394 & 0.009  & 0.432 & 0.036\\
$mean_m$ & 0.393 & 0.011  & 0.443 & 0.038\\
$w2s_d$ & 0.393 & 0.008  & 0.449 & 0.035\\
$w1s_d$ & 0.393 & 0.009  & 0.447 & 0.035\\
$w2s_m$ & 0.392 & 0.010  & 0.435 & 0.034\\
$w2_m$ & 0.391 & 0.011  & 0.452 & 0.030\\
$w2n_m$ & 0.391 & 0.012  & 0.439 & 0.034\\
\midrule
$GP_{linard}$ & 0.398 & 0.014 & 0.424 & 0.075\\
$GP_{linone}$ & 0.398 & 0.014 & 0.426 & 0.071\\
$SVM$ & 0.367 & 0.010 & 0.496 & 0.030\\
$SVMfs$ & 0.366 & 0.011 & 0.503 & 0.032\\
 \bottomrule
		\end{tabular}
		}
\end{table}

In Tables \ref{tab:ANEW_SWN_1} and \ref{tab:ANEW_SWN_3}, the results of the regression experiments over the ANEW dataset, using SWN\textsubscript{1} and SWN\textsubscript{3}, are presented. The results of the classification experiments over the GI dataset, using SWN\textsubscript{1} and SWN\textsubscript{3} are shown in Tables \ref{tab:GI_SWN_1} and \ref{tab:GI_SWN_3}. For the sake of interpretability, results are divided according to the main approaches: randoms, posterior-to-prior formulae, learning algorithms.  Note that for classification we report the generics $f$ and not the $f_m$ and $f_d$ variants. In fact, both versions always return the same classification answer (we are classifying according to the sign of $f$ result and not its strength). For the GPs, we report the two best configurations only.

\begin{table} [hb!]
	\caption{MAE results for regression using SWN\textsubscript{3}}
	\label{tab:ANEW_SWN_3}
	\centering
	{
		\begin{tabular}{@{}lrrrr@{}}
 \toprule
Approach  & MAE$_\mu$ & MAE$_\sigma$ & $\rho_\mu$ & $\rho_\sigma$ \\
\midrule
 $rnd$ & 0.652 & 0.026 & -0.002 & 0.123\\ 
$swnrnd_d$ & 0.404 & 0.013 & 0.395 & 0.018\\
$swnrnd_m$ & 0.402 & 0.010 & 0.399 & 0.036\\

\midrule
$max_m$ & 0.393 & 0.009  & 0.517 & 0.039\\
$fs_d$ & 0.382 & 0.008  & 0.544 & 0.029\\
$uniw_m$ & 0.382 & 0.015  & 0.490 & 0.049\\
$fs_m$ & 0.381 & 0.010  & 0.540 & 0.031\\
$median_m$ & 0.377 & 0.008  & 0.502 & 0.024\\
$uniw_d$ & 0.377 & 0.012  & 0.522 & 0.036\\
$median_d$ & 0.377 & 0.011  & 0.530 & 0.013\\
$uni$ & 0.376 & 0.010  & 0.493 & 0.030\\
$max_d$ & 0.372 & 0.011  & 0.549 & 0.028\\
$mean_d$ & 0.371 & 0.010 & 0.548 & 0.017\\
$w1sn_m$ & 0.371 & 0.011 & 0.527 & 0.040\\
$w2sn_m$ & 0.369 & 0.010 & 0.531 & 0.038\\
$w1_d$ & 0.368 & 0.010 & 0.567 & 0.020\\
$w2_d$ & 0.367 & 0.010 & 0.567 & 0.018\\
$mean_m$ & 0.367 & 0.010 & 0.527 & 0.029\\
$w1_m$ & 0.365 & 0.010 & 0.552 & 0.034\\
$w2sn_d$ & 0.364 & 0.011 & 0.554 & 0.026\\
$w1sn_d$ & 0.364 & 0.010 & 0.554 & 0.027\\
$w1s_m$ & 0.363 & 0.009 & 0.533 & 0.038\\
$w1n_d$ & 0.362 & 0.009 & 0.563 & 0.030\\
$w2s_d$ & 0.362 & 0.010 & 0.562 & 0.020\\
$w2_m$ & 0.362 & 0.010 & 0.554 & 0.032\\
$w1s_d$ & 0.362 & 0.009 & 0.561 & 0.022\\
$w1n_m$ & 0.362 & 0.007 & 0.549 & 0.045\\
$w2n_d$ & 0.361 & 0.010 & 0.563 & 0.030\\
$w2s_m$ & 0.360 & 0.009 & 0.540 & 0.035\\
$w2n_m$ & 0.359 & 0.009 & 0.551 & 0.043\\
\midrule
$GP_{linone}$ & 0.356 & 0.008 & 0.533 & 0.034\\
$GP_{linard}$ & 0.355 & 0.008 & 0.533 & 0.032\\
$SVM$ & 0.333 & 0.004 & 0.569 & 0.027 \\
$SVMfs$ & 0.333 & 0.003 & 0.568 & 0.027\\
\bottomrule
		\end{tabular}
		}

\end{table}

\begin{table} [ht]
	\caption{Accuracy results for classification using SWN\textsubscript{1}}
	\label{tab:GI_SWN_1}
	\centering
	{
		\begin{tabular}{@{}lrrrr@{}}
\toprule
Approach & Acc$_\mu$ & Acc$_\sigma$  & Kappa$_\mu$ & Kappa$_\sigma$ \\
\midrule
$rnd$ & 0.447 & 0.019 & 0.011 & 0.024\\ 
$majority\_class$ & 0.558 & 0.017 & 0.000 & 0.000\\
$swn\_rnd_m$ & 0.639 & 0.026 & 0.336 & 0.015\\
$swn\_rnd_d$ & 0.646 & 0.021 & 0.355 & 0.015\\
\midrule
$fs$ & 0.659 & 0.020  & 0.342 & 0.044\\
$uni$ & 0.684 & 0.017  & 0.364 & 0.035\\
$median$ & 0.686 & 0.022  & 0.374 & 0.047\\
$uniw$ & 0.702 & 0.019  & 0.395 & 0.033\\
$max$ & 0.710 & 0.022  & 0.410 & 0.038\\
$w1$ & 0.712 & 0.021  & 0.416 & 0.044\\
$w1n$ & 0.713 & 0.022  & 0.416 & 0.045\\
$w2n$ & 0.714 & 0.023  & 0.419 & 0.045\\
$w2$ & 0.715 & 0.021  & 0.420 & 0.047\\
$mean$ & 0.718 & 0.023  & 0.429 & 0.052\\
$w2s$ & 0.719 & 0.023  & 0.431 & 0.048\\
$w2sn$ & 0.719 & 0.023  & 0.431 & 0.048\\
$w1s$ & 0.719 & 0.023  & 0.432 & 0.048\\
$w1sn$ & 0.719 & 0.023  & 0.432 & 0.048\\
\midrule
$GP_{linard}^{lll}$ & 0.721 & 0.026 & 0.445 & 0.050\\
$GP_{linard}^{prl}$ & 0.722 & 0.025 & 0.447 & 0.048\\
$SVM$               & 0.733 & 0.021 & 0.458 & 0.042\\
$SVMfs$             & 0.743 & 0.021 & 0.474 & 0.047\\
\bottomrule
		\end{tabular}
		}
\end{table}

\begin{table} [ht]
	\caption{Accuracy results for classification using SWN\textsubscript{3}}
	\label{tab:GI_SWN_3}
	\centering
	{
	\begin{tabular}{@{}lrrrr@{}}
\toprule
Approach & Acc$_\mu$ & Acc$_\sigma$  & Kappa$_\mu$ & Kappa$_\sigma$ \\
\midrule
$rnd$ & 0.447 & 0.019 & 0.011 & 0.024\\ 
$majority\_class$ & 0.558 & 0.017 & 0.000 & 0.000\\
$swn\_rnd_d$ & 0.700 & 0.030 & 0.431 & 0.018\\
$swn\_rnd_m$ & 0.706 & 0.034 & 0.441 & 0.028\\
\midrule
$fs$ & 0.723 & 0.014  & 0.452 & 0.037\\
$median$ & 0.742 & 0.016  & 0.486 & 0.026\\
$uni$ & 0.750 & 0.015  & 0.492 & 0.029\\
$uniw$ & 0.762 & 0.023  & 0.504 & 0.027\\
$max$ & 0.769 & 0.019  & 0.518 & 0.027\\
$w2s$ & 0.777 & 0.017  & 0.531 & 0.017\\
$w2sn$ & 0.777 & 0.017  & 0.531 & 0.017\\
$w1s$ & 0.777 & 0.017  & 0.532 & 0.016\\
$w1sn$ & 0.777 & 0.017  & 0.532 & 0.016\\
$w1n$ & 0.780 & 0.021  & 0.544 & 0.027\\
$w2n$ & 0.780 & 0.022  & 0.545 & 0.026\\
$mean$ & 0.781 & 0.018  & 0.543 & 0.023\\
$w1$ & 0.781 & 0.021  & 0.547 & 0.027\\
$w2$ & 0.781 & 0.021  & 0.549 & 0.026\\
\midrule
$SVM$   & 0.779 & 0.016 & 0.553 & 0.033\\
$GPl$   & 0.779 & 0.018 & 0.558 & 0.035\\
$GPg$   & 0.781 & 0.018 & 0.562 & 0.036\\
$SVMfs$ & 0.792 & 0.014 & 0.577 & 0.029\\
\bottomrule
		\end{tabular}
		}
\end{table}

\subsection{Discussion} 
\label{sec:prior_discussion}

In this section, we sum up the main results of our analysis, providing an answer to the various questions we introduced at the beginning of the paper ({since results are largely consistent across the measurements both in regression and classification, in the following we will discuss MAE and accuracy only}):

\textbf{SentiWordNet improves over random.} One of the first things worth noting -- in Tables \ref{tab:ANEW_SWN_1}, \ref{tab:ANEW_SWN_3}, \ref{tab:GI_SWN_1} and \ref{tab:GI_SWN_3} -- is that the random approach (\emph{rnd}), as expected, is the worst performing metric, while all other approaches, based on SWN, have statistically significant improvements both for MAE and for accuracy ($p<0.001$). 

\textbf{SWN\textsubscript{3} is better than SWN\textsubscript{1}.} With respect to SWN\textsubscript{1}, using SWN\textsubscript{3} improves the results, both in regression ($MAE_\mu$ 0.398 vs. 0.366, $p < 0.001$) and classification ($accuracy_\mu$ 0.710 vs. 0.771, $p < 0.001$) tasks. Since many of the approaches described in the literature use SWN\textsubscript{1} their results should be revised and SWN\textsubscript{3} should be used as standard. This difference in performance can be partially explained by the fact that, even after pre-processing, for the ANEW dataset 137 lemma-PoS have all senses equal to 0 in SWN\textsubscript{1}, while in SWN\textsubscript{3} they are just 48. In the GI lexicon the same occurs for 223 lemma-PoS of SWN\textsubscript{1} and 69 of SWN\textsubscript{3}.

\textbf{Not all formulae are created equal.} The formulae described in Section \ref{sec:PriorF} have very different results, along a continuum. While inspecting every difference in performance is out of the scope of the present paper, we found that there is a strong difference between best and worst performing formulae both in regression and classification and these differences are all statistically significant ($p < 0.001$). 
Furthermore, the new formulae we introduced, based on the ``posterior polarities saliency'' hypothesis,  proved to be among the best performing in all experiments. This suggests that there is room for inspecting new formulae variants other than those already proposed in the literature.

\textbf{Selecting one sense is not a good choice.} On a side note, the approaches that rely on the polarity of a single sense (namely $fs$, $median$ and $max$) have similar results which do not differ significantly from $swnrnd$. 
These same approaches are also far from the best performing formulae: 
{the difference between the corresponding best performing formula and the single senses formulae is always significant in the various tables (at least $p < 0.01$).}
Among other things, this finding shows that taking the first sense of a lemma-PoS in some cases has no improvement over taking a random sense, and that in all cases it is one of the worst approaches with SWN. This is surprising since in many NLP tasks, such as word sense disambiguation, algorithms based on the most frequent sense represent a very strong baseline\footnote{In SemEval2010, only 5 participants out of 29 performed better than the most frequent threshold \cite{agirre2010SemEval}.}.  

\textbf{Learning improvements.} Combining the formulae in a learning framework {with our ensemble methods} further improves the results over the best performing formulae, both in regression ($MAE_\mu$ with SWN\textsubscript{1} 0.366 vs. 0.391, $p < 0.001$; $MAE_\mu$ with SWN\textsubscript{3} 0.333 vs. 0.359, $p < 0.001$) and in classification ($accuracy_\mu$ for SWN\textsubscript{1} is 0.743 vs. 0.719, $p < 0.001$; $accuracy_\mu$ for SWN\textsubscript{3} is 0.792 vs. 0.781, not significant $p = 0.07$). 
Another thing worth noting is that, in regression, GPs are outperformed by both versions of SVM ($p < 0.001$), see Tables \ref{tab:ANEW_SWN_1} and \ref{tab:ANEW_SWN_3}.  This is in contrast to the results presented in \cite{cohnmodelling}, where GPs on the single task are on average better than SVMs. In classification, GPs have similar performance to SVM without feature selection, and in some cases (see Table \ref{tab:GI_SWN_3}) even slightly better. 
In all our experiments, SVM with feature selection leads to the best performance. This is not surprising due to the high level of redundancy in the formulae scores. Interestingly, inspecting the most frequently selected features by $SVMfs$, we see that features from different groups are selected, and even the worst performing formulae can add information. 
This confirms the idea that viewing the same information from different perspectives (i.e. the posterior polarities provided by SWN combined in various ways {using ensemble methods}) can obtain better predictions.

 {To sum up, according to our results $SVMfs$ using SWN\textsubscript{3} outperforms all other methods for prior-polarity computation starting from SentiWordNet.}

\section{Error analysis}
\label{sec:error_analysis}

{
As a next step we wanted to understand why the learning algorithms perform better than the formulae. We inspected the errors of the best performing classifier ($SVMfs$) and of the best performing formula ($w2$) in the classification task, for a total of 652 misclassified lemma-PoS.
In particular, 67\% of the words are mislabeled by both methods, while $w2$ and $SVMfs$ mislabel 19\% and 14\% respectively.
A manual inspection shows that errors are mostly due to discrepancies between posterior polarity values in SWN and the gold label provided by GI. 
For example, \texttt{pretty\#a} has two senses, the first one being positive and the second negative.\\
To explore the nature of such discrepancies, we asked two annotators to inspect a subsample (50 elements) of the errors' dataset and classify whether the SWN values are correct or not, by looking at each lemma-PoS-sense value and comparing it with the synset gloss. This analysis revealed that 76\% of the errors are determined by incorrect values in SWN (with a good annotators agreement, Cohen's kappa = .75). For example, the synset \texttt{overjoyed\#a} has only one sense, with \texttt{Pos(s)} = 0,  \texttt{Neg(s)} = 0.75, and this means that both $SVMfs$ and $w2$ rate the word as negative even though it is positive. On the other hand, the second sense of \texttt{pretty\#a} refers to the ironical use of the word, so its negative value is fine. Given such discrepancies, we identified how they affect $w2$ and $SVMfs$:\\
\emph{(i)} when there is an error in a lemma-PoS with only one sense (e.g. the aforementioned \texttt{overjoyed\#a}), or errors are distributed over all senses, both methods will fail to find the correct label. This is the case for about 28\% of the misclassified words in our dataset.\\
\emph{(ii)} When the first sense has a posterior polarity different from the gold label, $w2$ usually gives an incorrect label, as the first sense is weighted much more than all the others. Instead, $SVMfs$ can still find the correct prior since it is less sensitive to noisy data, and it considers all the other senses in a more grounded way. For example, \texttt{wickedness\#a} has a positive value in the first sense (0.75) and mostly negative values for the remaining 4 senses, but it is nevertheless classified as positive by $w2$, while it is correctly labelled as negative by $SVMfs$.\\ 
\emph{(iii)} When the first sense has the same polarity as the gold label, and most other senses have the opposite sign, $SVMfs$ usually assigns the wrong label, while $w2$ does not. Albeit less common this happens for words such as \texttt{confident\#a},  whose \texttt{Pos(s)} are +0.375, +0.125 and +0.125, while the \texttt{Neg(s)} are 0, -0.375 and -0.625. The classifier is affected by the strong negativity of the last two senses and incorrectly classifies it as negative.
}{

}

\begin{figure}[h!] 
\begin{center}
\includegraphics[width=\columnwidth]{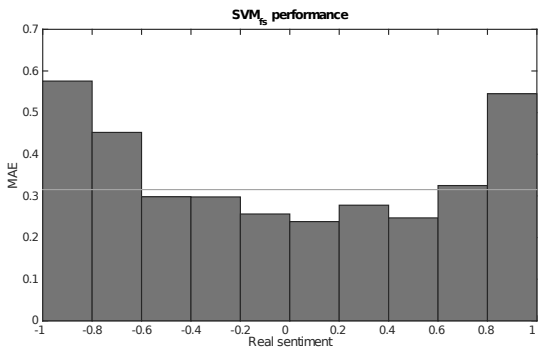}
\caption{MAE values per bins over ANEW dataset.} 
\label{fig:error_bins} 
\end{center} 
\end{figure}

We also did a similar error analysis for the regression task, by defining as an error a MAE that is greater than 2 standard deviations of the overall MAE distribution. The results are in line with the previous analysis, in particular the fact that $SVMfs$ can recover from errors or incoherent values in SWN scores better than simple formulae. 
{Finally, in Figure \ref{fig:error_bins} we report the MAE of $SVMfs$ according to ANEW bins (the horizontal line being the average MAE on the whole dataset), in order to understand how the errors are distributed. On average, our method is less precise on extreme values, where the number of training samples from ANEW is lower.} 

\section{\emph{SentiWords}} 
\label{sec:SentiWords}
{In the previous sections we have shown how an ensemble method ($SVMfs$) can be used to calculate more accurate prior polarities, starting from the posterior polarities scores of SWN\textsubscript{3}. We used these results to create \emph{SentiWords}, a lexicon that maximizes both precision and coverage. To obtain this result, we trained our classifier on a larger dataset, the 13,915 entries of \emph{Warr} \cite{Warriner2013norms}, and used it to annotate all the lemma-PoS of SWN$_3$.}

In particular, we processed \emph{Warr} as we did with ANEW (see Section \ref{sec:exp}). 
This way, we obtained a list of 18,154 lemma-PoS, each one associated with the valence score given by human annotators, paired with  the scores given by the formulae selected with randomized lasso as features. We used this as training, to create a  more precise $SVMfs$ regression model.
All the lemma-PoS of SWN\textsubscript{3} for which we had at least one non-0 value (roughly 40,000) were thus scored using the $SVMfs$  model. Finally we merged this list with \emph{Warr} to obtain  \emph{SentiWords}. 
In a similar way we also created, for our experiments, \emph{SentiWords\textsubscript{bin}}, using the complete list of GI as a training set.

%%%%%%%%%%%%%%%%%%%%%%%
%%%%%%%%%%%%%%%%%%%%%%%
%%%%%%%%%%%%%%%%%%%%%%%
\section{Prior polarities and Sentiment Analysis} 
\label{sec:2_exp}
%%%%%%%%%%%%%%%%%%%%%%%
%%%%%%%%%%%%%%%%%%%%%%%
%%%%%%%%%%%%%%%%%%%%%%%

To validate the improvement we can obtain in sentiment analysis with \emph{SentiWords} over the single metrics and over other widely used handmade lexica like ANEW -- that are more precise but have a much smaller coverage -- we ran an extensive series of experiments. In these experiments we considered 2 datasets of sentences annotated both with sentiment values (ranging from -1 to 1) and sentiment labels (\texttt{NEGATIVE} or \texttt{POSITIVE}).

As a comparison with \emph{SentiWords}, we considered also 4 human-annotated lexica (ANEW, \emph{Warr} and \emph{Stanf} as gold standards for regression, the same for classification but with GI instead of ANEW) to test the importance of coverage and precision of our newly built lists. In particular: 

\begin{itemize}
\item ANEW represents a gold standard with low coverage on a continuous scale.
\item GI represents a gold standard with low coverage in a binary format.
\item \emph{Warr} represents a gold standard with high coverage (at present the highest coverage available for prior polarities). 
\item \emph{Stanf} represents a gold standard with high coverage but with "posterior-polarities".\footnote{The \emph{Stanf} lexicon is not available \emph{per se}, we created it by extracting all the single words present in the Stanford Sentiment Treebank (see section \ref{subsec:stb} for a description of the dataset), with their manually annotated affective score.} 
\end{itemize}

To be able to compare the results of the experiments, all these lexica were transformed to a \texttt{lemma\#PoS} format as described in Section \ref{sec:exp}. The final size of each lexicon is reported in Table \ref{tab:lexica_dimensions}.

\begin{table} [ht] 	
	\caption{Lexica sizes} 	
	\label{tab:lexica_dimensions} 
	\centering
			\begin{tabular}{lr} 						
\toprule
Lexicon & lemma\#PoS entries \\
\midrule
ANEW & 1,483 \\
GI & 3,041 \\
Stanf & 15,223 \\
Warr & 18,005 \\
SentiWords & 155,286 \\
\bottomrule
\end{tabular} 		

\end{table} 

\subsection{Datasets}
\label{sec:datasets}

To assess how well the use of prior polarities performs on the specific task of text based sentiment analysis, we tested our resource and the gold standards lexica on two different datasets, that represents different form of language use, i.e. news headlines, with simplified syntax and lexicon, and sentences extracted from movie reviews, with normal language use. These two datasets are used in regression and classification tasks. 

\subsubsection{SemEval}

The public dataset provided for the SemEval2007 task on `Affective Text'~\cite{strapparava2007SemEval} is focused on emotion recognition in 1,000 news headlines, 
both in regression and classification settings. Headlines typically consist of a few words and are often written with the 
intention of `provoking' emotions to attract the readers' attention. An example of a headline from the dataset is the following: ``\emph{Iraq car bombings kill 22 People, wound more than 60}". For the regression task the value provided is -0.98, 
while for the classification task the label provided is \texttt{NEGATIVE}.

This  dataset (which will be referred to as SemEval henceforth) is  of interest to us since the `compositional' problem is less prominent given the simplified 
syntax of news headlines, containing, for example, fewer adverbs (like negations or intensifiers) than normal sentences~\cite{turchi2012onts}. 
Each headline of the dataset was lemmatized and PoS tagged, keeping only those lemma-PoS that have a PoS mappable to WordNet. The average length of headlines is 7.21 words, (5.4 lemma-PoS). 
Only one headline contained just words not present in \emph{SentiWords}, further indicating the high-coverage nature of our resource. 

In Table \ref{tab:SemEval_coverage} we report the coverage of the Sentiment Lexica  on the SemEval dataset (i.e. percentage of words in the sentences recognized by the lexica). 
Of particular interest here is the fact that, since \emph{Warr} was built starting from the most commonly used English words, it grants a high coverage -- higher than the \emph{Stanf} Lexicon that has more entries but was built starting from a specific dataset. On the contrary, ANEW and GI show a very poor coverage and for almost half of the sentences there was no sentiment word recognized.

\begin{table} [ht] 
	\caption{Lexica coverage for the SemEval dataset} 	
	\label{tab:SemEval_coverage} 
	\centering	
			\begin{tabular}{@{}lr@{}} 						
\toprule
Lexicon	& Coverage\\ 
\midrule
	GI & 0.08\\
	ANEW & 0.13 \\
	Warr & 0.69 \\
	Stanf &  0.66 \\
	SentiWords &  0.89 \\
\bottomrule
			\end{tabular} 			 	

\end{table} 

\subsubsection{Sentiment Treebank}
\label{subsec:stb}

The Stanford Sentiment Treebank (STB) is a corpus with fully labelled parse trees, that allows for a complete analysis of the compositional effects of sentiment in language \cite{socher2013recursive}. The corpus is based on the dataset introduced in \cite{Pang+Lee:05a} and consists of 11,855 single sentences extracted from movie reviews. It was parsed with the Stanford parser \cite{klein2003accurate} and includes a total of 215,154 unique phrases from those parse trees, each annotated by 3 human judges (using Mechanical Turk).

An example of a movie review sentence  is: ``\emph{One of the finest, most humane and important Holocaust movies ever made.}". For the regression task the value provided is +0.97, 
while for the classification task the label provided is \texttt{POSITIVE}.

For our experiments we took the 11,855 sentences of the STB dataset and lemmatized and PoS tagged all the words, keeping only those lemma-PoS that had a PoS mappable on WordNet, as was done with the SemEval dataset. The average length of a sentence in STB is 20.4 words (11 lemma-PoS). 

This  dataset is  somehow complementary to the previous one, since here the syntax is not simplified and represents ``natural" language use. 

In Table \ref{tab:coverage_STB} we report the coverage of our Sentiment Lexica on the STB dataset. 
Results are similar to the previous case, with ANEW and GI showing a very poor coverage: for about 40\% of the sentences there was no sentiment word recognized. Note that \emph{Stanf} has the same coverage as \emph{SentiWords} -- even if it is much smaller -- since it was built starting from the words present in the STB itself and discarding those that could not be aligned with SWN entries.    

\begin{table} [ht] 	
	\caption{Lexica coverage for the STB dataset} 	
	\label{tab:coverage_STB} 
	\centering
			\begin{tabular}{@{}lr@{}} 						
\toprule
Lexicon	& Coverage\\ 
\midrule
GI & 0.10\\
ANEW & 0.11 \\
Warr & 0.65 \\
Stanf & 0.86 \\
SentiWords & 0.86\\
\bottomrule
			\end{tabular} 		 	 	
\end{table}

\section{Sentiment Analysis Experiments}
The experiments presented in this section are performed both by using sentences as present in the datasets and by filtering stop words from them. 
The rationale for this choice is given by the fact that prior-polarity scores can also be given to words that, for the task of text-based sentiment analysis, are not ``relevant", like auxiliary verbs, biasing the results. Still, it is not an error \emph{per se}, to give a score to such stop words: if people perceive that they convey an affective meaning when taken in isolation, this information can be very useful for other sentiment-related tasks. Going back to the example of naming described in the introduction, let us consider the paradigmatic example of perfumes, that tend to use evocative names -- since their smell cannot be "shown" in advertisement: we have  "\emph{Must}" from Cartier, or "\emph{Be}" from CalvinKlein, which are both auxiliary verbs. Both examples have a positive score (usually in advertising we want a positive feeling associated with the brand) and according to \emph{Warr}: \texttt{be\#v} +0.300, \texttt{must\#v} +0.113\footnote{A similar example can be drawn for downtoners or intensifiers (i.e. words -- such as \emph{slightly}, \emph{somewhat} or \emph{very}, \emph{completely} -- that decreases or increase the effect of a modified item). These words are usually adverbs or adjectives (like \emph{small} or \emph{big}) and while for the task of sentiment analysis they need to be considered as special linguistic objects for compositional purposes, when taken in isolation they can have their own affective score. Consider the vodka brand ``Absolut" (pronounced as the intensifier adjective \texttt{absolute\#a} with a positive score of +0.108.)}.

To have a fair comparison among the lexica, we rely on a standard list of stop words (the MySQL stopword list for MyISAM search indexes, consisting of 543 tokens) rather than creating one specifically tailored to our datasets or task. Stop words in this list are thus removed from the datasets in the corresponding experimental setting.

Furthermore, to have a fair comparison of resources performance (i.e. without any syntactic or compositional reasoning that can boost the performance) we used a na\"{\i}ve approach that averages over all the word scores in a sentence, similar, for example, to the approaches used in \cite{strapparava2008learning} and \cite{staiano2014depechemood}. In particular for the regression experiments we use the "average" of the corresponding affective scores -- obtained from the lexicon under inspection -- of all lemma-PoS recognized in the text{, so the sentence ``Families celebrate return of sons'' \emph{(i)} gets PoS-tagged to ``\texttt{family\#n celebrate\#v return\#n son\#n}, \emph{(ii)} for each resource to test, the word scores are found and averaged. For example, for \emph{SentiWords} the result will be $(0.562+0.710+0.237+0.477)/4=0.497$, while for \emph{Stanf} it will be $[0.333+0.667+(-0.055)+0]/4=0.236$.}
In classification experiments a majority vote over the single words {is used to predict sentiment (e.g. ``Massive mud traps dozens of families'' will become \texttt{massive\#a mud\#n trap\#v family\#n}, which through \emph{SentiWords\textsubscript{bin}} gets assigned the value $0 + (-1) + (-1) + 1 = -1$, i.e. a negative label).}

For the sake of conciseness in the following we report only the result of the best and the worst performing prior formulae -- using SWN\textsubscript{3} -- for each experiment ($f_{best}$ and $f_{worst}$ respectively). In general, the results for these formulae are consistent with the experiments carried out on prior-polarity computations, discussed in Section \ref{sec:prior_discussion} (e.g. $fs$ being one of the worst approaches also in sentiment analysis).  Moreover, to test the importance of sample size for learning prior polarities, together with \emph{SentiWords} results, we also report  the regression results of the best learning model that was built using ANEW ($SVMfs$).  
{To give a comparison we also report, separately, the results obtained by CLaC \cite{andreevskaia2007clac}, the best performing system at SemEval 2007 (indicated in the tables with SemEval\textsubscript{best}). CLaC is an unsupervised system, i.e. without prior knowledge of this dataset. To detect headline sentiment, it uses a list of ``sentiment-bearing unigrams'', constructed by expanding a small set of human-annotated positive and negative words using WordNet synonymy and antonymy relations, and adding G.I. Positive and Negative words too. In total, 10,809 sentiment-bearing words with different PoS are used. CLaC also uses a list of 490 valence shifters (e.g. negations, intensifiers, etc.) and rules for defining the scope and the results of the combination of sentiment-bearing words and value shifters.}

In Tables \ref{tab:SemEval} and \ref{tab:STANFORD_regression}, the results of the regression experiments -- over the SemEval and the STB datasets respectively -- are presented.
In this case we chose to use Pearson's correlation coefficient instead of MAE since (i) it is the official measurement of SemEval2007 and (ii) it is not sensitive to data scaling/normalization, unlike MAE, so we can directly compare the averages returned by our na\"{\i}ve approach with the gold standard scores.

\begin{table} [ht]
	\caption{Correlation results for regression on SemEval.\protect\footnotemark}
	\label{tab:SemEval}
	\centering
		\begin{tabular}{@{}lc@{}}
\toprule
Lexicon & $\rho$ \\
\midrule
$f_{worst}$ & 0.253\\
ANEW & 0.270 \\
$f_{best}$ & 0.382 \\
$SVMfs$ & 0.410 \\
Stanf & 0.427\\
Warr & 0.567 \\
SentiWords & 0.570 \\
\midrule
Lexicon (removing stop words) & $\rho$\\
\midrule
$f_{worst}$ & 0.257\\
ANEW & 0.268 \\
$f_{best}$ & 0.373 \\
$SVMfs$ & 0.387 \\
Stanf & 0.428\\
Warr & 0.555 \\
SentiWords & 0.557 \\
\midrule
{SemEval\textsubscript{best}} & {0.477} \\
\bottomrule
\end{tabular}

\end{table}

\begin{table} [ht]
	\caption{Correlation results for regression on STB}
	\label{tab:STANFORD_regression}
	\centering
		\begin{tabular}{@{}lc@{}}
\toprule
Lexicon & $\rho$\\
\midrule
ANEW & 0.175 \\
$f_{worst}$ & 0.268\\
$SVMfs$ & 0.321 \\
$f_{best}$ & 0.328 \\
Warr & 0.359 \\
SentiWords & 0.377 \\
Stanf	& 0.495 \\
\midrule
Lexicon (removing stop words) & $\rho$\\
\midrule
ANEW & 0.177 \\
$f_{worst}$ & 0.284\\
$f_{best}$ & 0.335 \\
$SVMfs$ & 0.350 \\
Warr & 0.384 \\
SentiWords & 0.402 \\     
Stanf &	0.496 \\
\bottomrule
\hline
\end{tabular}

\end{table}

\footnotetext{In the following tables, we use these abbreviations: $f_{worst}$ is the worst performing $SWN_3$ formula, $f_{best}$ is the best performing $SWN_3$ formula, $SVMfs$ refers to the SVM trained on ANEW and SemEval\textsubscript{best} is the best performing system at SemEval 2007.}

The results of the classification experiments over the SemEval and STB datasets are shown in Tables \ref{tab:SemEval_classification} and \ref{tab:STANFORD_classification} respectively. 

In the SemEval2007 task, sentences in the range $[-1, -0.5]$ were considered negative, while those in the range $[0.5, 1]$ were labelled as positive. We took this division for our classification experiments, discarding neutral sentences (i.e. those ranging from -0.5 to 0.5), thus obtaining 410 entries with a binary polarity score, 62\% of which were negative and the remaining 38\% positive. 
We applied to the STB dataset the same ``binarization'' that we used for SemEval, thus filtering out neutral sentences. The final dataset for the classification experiments consisted of 5,365 sentences, of which 54\% were positive and 46\% were negative. We also ran the same experiments on a dataset created with stricter positivity and negativity threshold (i.e., considering the sentences that fall in the range $[-1, -0.2]$ negative and those which fall in the range $[0.2, 1.0]$ positive, as suggested by the STB instruction file). Since the results are consistent for both datasets, we present those relating to the dataset created using the SemEval technique.
In the following section, to check if there is a statistically significant difference in the results, we used Fisher's z-transformation for the correlations, and the approximate randomization test for classification experiments.

\begin{table} [ht]
	\caption{Accuracy results for classification on SemEval}
	\label{tab:SemEval_classification}
	\centering
		\begin{tabular}{@{}lr@{}}
\toprule  
Lexicon & Accuracy \\
\midrule
GI                  				& 0.317 \\
Stanf					& 0.529 \\
$f_{worst}$	  			& 0.448\\
$f_{best}$					& 0.571 \\
SentiWords\textsubscript{bin} 	& 0.581 \\
\midrule
Lexicon (removing stop words) & Accuracy \\
\midrule
GI 						& 0.317 \\
Stanf					& 0.556 \\
$f_{worst}$ 				& 0.424\\
$f_{best}$					& 0.586 \\
SentiWords\textsubscript{bin} 	& 0.602 \\
\midrule
{SemEval\textsubscript{best}} & {0.551} \\
\bottomrule
		\end{tabular}
\end{table}

\begin{table} [ht]
	\caption{Accuracy results for classification on STB}
	\label{tab:STANFORD_classification}
	\centering
		\begin{tabular}{@{}lr@{}}
\toprule  
Lexicon & Accuracy \\
\midrule
GI 						& 0.416 \\
$f_{worst}$ 				& 0.536	\\
$f_{best}$					& 0.567 \\
SentiWords\textsubscript{bin}  	& 0.586 \\
Stanf 					& 0.604 \\
\midrule
Lexicon (removing stop words) & Accuracy \\
\midrule
GI 						& 0.414 \\
$f_{worst}$	 			& 0.528\\
$f_{best}$					& 0.576 \\
SentiWords\textsubscript{bin} 	& 0.595 \\
Stanf				& 0.633 \\
\bottomrule
\end{tabular}
\end{table}

\subsection{Discussion} 
\label{Discussion_Sentiment_Analysis} 

In this section, we sum up the main results of our experiments, providing an answer to the questions we introduced at the beginning of the paper: 

\textbf{Size matters (learning)}. The use of \emph{Warr} gives a boost in performance to \emph{SentiWords} compared to the scores returned by $SVMfs$, which is based on ANEW learning sample ($\rho$ values are more than double in both SemEval and STB datasets, both with and without stop words, $p < 0.001$). In fact both resources cover the whole SWN list of  155,000 lemma-PoS but since \emph{SentiWords} was built starting from a resource (\emph{Warr}) that contains 12 times the examples of ANEW, we can conclude that this doubling in performance is given by the initial learning sample size.  

\textbf{Size matters (coverage)}. ANEW is a very precise lexicon but, due to the small size, its coverage is very low, with many cases of ``undecidable'' sentences (i.e. sentences for which there are no words in the lexicon). This leads to poor performance when compared to \emph{SentiWords} ($\rho$ values are more than double in all regression experiments, $p < 0.001$), being in some cases even worse than the worst SWN\_formula, see Table \ref{tab:STANFORD_regression}. In classification, the same holds for GI versus \emph{SentiWords\textsubscript{bin}} on both datasets ($p < 0.001$).

\textbf{Priors, less precise but portable}. The comparison with the \emph{Stanf} lexicon (which is ``over-fitted" on the STB dataset) shows that using posterior polarities can yield better results when used on specific datasets, see Table \ref{tab:STANFORD_regression} and \ref{tab:STANFORD_classification}. Still, when used on different datasets and in different scenarios, the performance drastically decreases, see Table \ref{tab:SemEval} and \ref{tab:SemEval_classification}. The average correlation on the datasets is higher for \emph{SentiWords} as compared to \emph{Stanf} ($\rho_\mu$
 0.480 vs. 0.462 in the stop words setting). In classification the difference is less marked ($accuracy_\mu$ 0.599 vs 0.595), but while \emph{SentiWords\textsubscript{bin}} performance is almost identical across datasets, \emph{Stanf} has a drop on the SemEval dataset. We can reasonably conclude that, if we were to consider additional datasets and domains, the difference between \emph{SentiWords} and posterior lexica (i.e. \emph{Stanf}) would increase.
 
\textbf{Stop words}. The removal of stop words significantly increases the performance of our Lexica {in the regression task}, especially on the STB dataset ($p < 0.05$ for $SVMfs$, \emph{Warr} and \emph{SentiWords}). While the SemEval headlines use a simplified language (also with less stop words), the movie reviews use a plain language. In particular, 12\% of lemma-PoS recognized by \emph{SentiWords} were discarded from SemEval because they were in the stop words list,  while in STB this number was more than double accounting for 26\% of the  \emph{SentiWords} lemma-PoS discarded.
As we could have expected, \emph{Stanf} is less sensitive to stop words removal since it is composed of posterior polarities.

\textbf{Precision vs. Coverage: the Losers}. SWN formulae only beat ANEW in regression (and $SVMfs$ trained on ANEW beats the formulae on average, consistently with results in section \ref{sec:prior_discussion}). The same holds for SWN formulae and GI in classification. That is to say: SWN metrics are better because of the high coverage compared to the two gold-prior lexica, but their precision is very low compared to \emph{Warr} and \emph{SentiWords}.

\textbf{Precision vs. Coverage: the Winners}. \emph{Warr} and \emph{SentiWords} performance is comparable on SemEval headlines, even if the \emph{Warr} lexicon is much smaller.
On STB, \emph{SentiWords} performs better in the stop words removal setting ($\rho$ 0.402 vs. 0.384, $p = 0.05$) and slightly better without removing them. 
This makes sense: on SemEval -- that has a simplified language, as news headlines use only very frequent terms -- the way \emph{Warr} was built (i.e. considering the most frequent words in English) grants that only few words are not covered, so performance is comparable. On \emph{Stanf}, where there are more words not covered by \emph{Warr}, what we learnt using ML for other words helps to improve performance. 
In general, \emph{SentiWords} is of help in any dataset for which \emph{Warr} has lower coverage.

{Finally, for the sake of comparison, we consider also SemEval\textsubscript{best} (the best performing system at SemEval 2007).
In our experiments, this system scored worse than \emph{Warr} and \emph{SentiWords} in regression, and worse than $f_{best}$ and \emph{SentiWords} in classification. 
These results give further evidence of the importance of a precise and high-coverage lexicon, in fact SemEval\textsubscript{best} uses elaborated compositional strategies but with a poor lexicon as compared to \emph{SentiWords}.}

{To sum up: according to our results, and to the best of our knowledge, \emph{SentiWords} represents a new state-of-the-art prior-polarity lexicon for sentiment analysis. It outperforms other SWN posterior-to-prior formulae and handmade lexica thanks to its wide coverage and to the \emph{Warr} lexicon it was built on.}

\section{Conclusions}
\label{sec:conclusions}

 In this paper, we have presented a study on Prior Polarity lexica for sentiment analysis.  While manually annotated lexica provide a high precision but lack of coverage,  automatic derivation from pre-existing knowledge guarantees high coverage at the cost of a lower precision. Starting from the experience of automatic derivation of prior polarities  from the SentiWordNet resource,  we used an ensemble learning framework  that -- taking advantage of manually built lexica -- is able to better predict the prior value of unseen words. We concluded by demonstrating that it is possible to use this technique to create a resource (\emph{SentiWords}) with a very high coverage and a good precision. Using our lexicon in sentiment analysis tasks,  we were able to  outperform both the single metrics derived from SentiWordNet and popular manually annotated sentiment lexica.

\section*{Acknowledgments}
\addcontentsline{toc}{section}{Acknowledgments}
The authors thank Jos\'{e} Camargo De Souza for his help with feature selection. 
This work has been partially supported by the Trento RISE PerTe project.

\bibliographystyle{IEEEtran}
\bibliography{IEEEabrv,prior_TAC}

\medskip

\section*{Biographies}
%\begin{IEEEbiography}[{\includegraphics[height=1.25in]{biopic_gatti.jpg}}]{Lorenzo Gatti}
\textbf{Lorenzo Gatti} 
is currently a Ph.D. student in the Department of Information Engineering and Computer Science at the University of Trento, Italy, funded by Fondazione Bruno Kessler (FBK-Irst). He received his master's degree in Cognitive Sciences from the University of Trento, with a specialization in Language and Multimodal Interaction. His current research interests include sentiment analysis, automatic humor generation and persuasive language. 
%\end{IEEEbiography}
\smallskip{}
%\begin{IEEEbiography}[{\includegraphics[height=1.25in]{biopic_guerini.jpg}}]{Marco Guerini}

\textbf{Marco Guerini} 
is a researcher in Computational Linguistics, focusing on persuasive communication, sentiment analysis and social media.
Currently he is working at FBK-Irst and previously at Trento-Rise, node of the European Institute of Technology after several years of research at FBK-Irst.
He graduated in Philosophy and holds a Ph.D. in Information and Communication Technologies since 2006. In 2011 his activities have been partially funded by a Google Research Award.
He is author of several scientific publications -- published in top-level conference proceedings and international
journals -- and program committee member at international conferences.
Since 2008 he also started working as a technology consultant for start-ups and large companies.
%\end{IEEEbiography}
\smallskip{}
%\begin{IEEEbiography}[{\includegraphics[height=1.25in]{biopic_turchi.jpg}}]{Marco Turchi}

\textbf{Marco Turchi} 
is currently tenure track researcher at Fondazione Bruno Kessler (FBK-Irst) in Italy. He received his Ph.D. in Computer Science from the University of Siena, Italy in 2006. Before joining FBK, he worked at the European Commission Joint Research Centre, Italy, at the University of Bristol, at the Xerox Research Centre Europe, and at Yahoo Research Labs. His current research interests include sentiment analysis and machine learning techniques applied to machine translation. He co-authored several scientific publications at top conferences and journals, and served as a reviewer for international journals, conferences, and workshops.
%\end{IEEEbiography}

\end{document}